# Multiview Hessian regularized logistic regression for action recognition

W. Liu, H. Liu, D. Tao*, Y. Wang, Ke Lu

China University of Petroleum (East China)

Abstract:

With the rapid development of social media sharing, people often need to manage the growing volume of multimedia data such as large scale video classification and annotation, especially to organize those videos containing human activities. Recently, manifold regularized semi-supervised learning (SSL), which explores the intrinsic data probability distribution and then improves the generalization ability with only a small number of labeled data, has emerged as a promising paradigm for semiautomatic video classification. In addition, human action videos often have multi-modal content and different representations. To tackle the above problems, in this paper we propose multiview Hessian regularized logistic regression (mHLR) for human action recognition. Compared with existing work, the advantages of mHLR lie in three folds: (1) mHLR combines multiple Hessian regularization, each of which obtained from a particular representation of instance, to leverage the exploring of local geometry; (2) mHLR naturally handle multi-view instances with multiple representations; (3) mHLR employs a smooth loss function and then can be effectively optimized. We carefully conduct extensive experiments on the unstructured social activity attribute (USAA) dataset and the experimental results demonstrate the effectiveness of the proposed multiview Hessian regularized logistic regression for human action recognition.

Highlights:

(1) mHLR combines multiple Hessian regularization to leverage the exploring of local geometry;

(2) mHLR can handle multi-view instances with multiple representations;

(3) mHLR employs a smooth loss function and then can be effectively optimized.

## 1. Introduction

The prodigious development of devices capable of digital media capture and internet technology makes vast volumes of multimedia data be uploaded and shared on the internet. It is an immediate need for effective techniques that can help manage this massive amount of data including video annotation, multimedia retrieval, multimedia understanding. Due to the expensive labor of labeling a large number of media data for model learning, semi-supervised learning (SSL) has attracted much attention since it can improve the generalization ability of a learning model by exploiting both a small number of labeled data and a large number of unlabeled data.

The most traditional class of SSL methods is based on the manifold regularization [1] which assumes that examples with similar features tend to have similar label attributes. Manifold regularization aims to explore the local geometry of data distribution using a penalty term to steer the regression function along the potential manifold. Representative manifold regularization includes locally linear embedding (LLE) [2] , ISOMAP [3] , local tangent space alignment (LSTA) [4] , Laplacian regularization (LR) [5] , and Hessian regularization (HR) [6] . Particularly, LLE uses linear coefficients to represent the local geometry. ISOMAP is a variant of MDS [7] which preserves global geodesic distances of all pairs of measurements. LSTA represents the local geometry by exploiting the local tangent information. LR develops approximate representation on Laplacian graph. HR obtains the presentation by estimating the Hessian over neighborhood. Comparing with other manifold regularization methods, Hessian has richer nullspace and drivers the learned function varying linearly along the underlying manifold [6] [8] [9] [21] . Thus Hessian regularization is preferable for encoding the local geometry and then boost the SSL performance.

On the other hand, multimedia is naturally represented by multiview features including spatial and time information. Different views character different data properties and features for different views are often complimentary to on another. Thus multiview features can provide more characteristics and significantly improve the learning performance. Popular multiview learning algorithms include co-training [10] , multiple kernel learning (MKL) [11] [12] [13] [14] , and graph ensemble learning [15] [16] [17] [18] . Co-training based algorithms learn from two different views to exploit unlabeled data to improve the classification performance. MKL learns a kernel machine from multiple Gram kernel matrices which are built from different view features. Graph ensemble based methods explore the complementary properties of different views by integrating multiple graphs which encode the local geometry of different views.

Based on the concerns above, in this paper we present multiview Hessian regularized logistic regression (mHLR) for action recognition. Significantly, mHLR seemly integrates multiview features and Hessian regularizations constructed from different views. The merits of mHLR lie in three folds: (1) mHLR combines multiple Hessian regularization to drive the leaned function varying linearly along the underlying manifold and then leverages the exploring of local geometry; (2) mHLR naturally handle multi-view instances with multiple representations to boost action recognition; (3) mHLR employs a smooth loss function and then can be effectively optimized. We conduct action recognition experiments on the unstructured social activity attribute (USAA) dataset [19] [20] . The experimental results show the effectiveness of mHLR by comparing with the baseline algorithms.

The rest of this paper is assigned as follows. Section 2 presents the proposed multiview Hessian regularized logistic regression. Section 3 details the optimization algorithm of mHLR. Section 4 discusses the experimental results on USAA dataset, followed by the conclusion in section 5.

2. Multiview Hessian regularized logistic regression

In multiview Hessian regularized logistic regression, we are given $l$ labeled examples $\mathcal{L} = \{(x_i^1, x_i^2, \cdots, x_i^V, y_i)\}_{i=1}^{l}$ and $u$ unlabeled examples $\mathcal{U} = \{(x_i^1, x_i^2, \cdots, x_i^V)\}_{i=l+1}^{l+u}$, where $V$ is the number of views, $x_i^k \in \mathcal{X}^k$ for $k \in \{1, 2, \cdots, V\}$ is the feature vector of the $k^{th}$ view of the $i^{th}$ example, and $y_i \in \{\pm 1\}$ is the label of example $x_i$. In the rest section of this paper, we use $x_i = \{x_i^1, x_i^2, \cdots, x_i^V\}$ to represent the $i^{th}$ example and $x^k$ to represent the $k^{th}$ view feature. Both labeled and unlabeled examples are drawn from the underlying manifold $\mathcal{M}$. Typically, $l \ll u$ and the goal is to predict the labels of unseen examples.

By incorporating a manifold regularization term to control the complexity of the learning function, the mHLR can be written as the following optimization problem:

$$\min_{f \in \mathcal{H}_K} \frac{1}{l} \sum_{i=1}^{l} \varphi(f, x_i, y_i) + \gamma_K \|f\|_K^2 + \gamma_I \|f\|_I^2 \quad (1)$$

where $\varphi(\cdot)$ is the logistic loss function, $\|f\|_K^2$ penalizes the classifier complexity in an appropriate reproducing kernel Hilbert space (RKHS) $\mathcal{H}_K$, $\|f\|_I^2$ controls $f$ along the compact manifold $\mathcal{M}$, and $\gamma_K$ and $\gamma_I$ are parameters which balance the loss function and regularization terms $\|f\|_K^2$ and $\|f\|_I^2$ respectively.

As mentioned above, examples are represented by multiview features. Then the proposed mHLR integrates multiple kernel learning and ensemble graph Hessian learning.

Suppose $K^k, k = 1, 2, \cdots, V$ is a valid (symmetric, positive definite) kernel on the

$k^{th}$ view, and then we define the multiview kernel as follows:

$$K = \sum_{k=1}^{V} \theta^k K^k, s.t. \sum_{k=1}^{V} \theta^k = 1, \theta^k \geq 0, k = 1,2,\cdots,V.$$

Suppose $K^k = \langle \phi_k(\omega^k), \phi_k(\mu^k)\rangle : X \times X \rightarrow R$, then $\theta^k K^k = \langle \sqrt{\theta^k}\phi_k(\omega^k), \sqrt{\theta^k}\phi_k(\mu^k)\rangle$.

Then we have

$$K(\omega,\mu) = \sum_{k=1}^{V} \theta^k K^k(\omega^k,\mu^k)$$

$$= \sum_{k=1}^{V} \langle \sqrt{\theta^k}\phi_k(\omega^k), \sqrt{\theta^k}\phi_k(\mu^k)\rangle$$

$$= \langle [\sqrt{\theta^1}\phi_k(\omega^1)\cdots\sqrt{\theta^k}\phi_k(\omega^k)\cdots\sqrt{\theta^V}\phi_k(\omega^V)], [\sqrt{\theta^1}\phi_k(\mu^1)\cdots\sqrt{\theta^k}\phi_k(\mu^k)\cdots\sqrt{\theta^V}\phi_k(\mu^V)]\rangle.$$

Therefore, the multiview kernel $K$ is also a valid (symmetric, positive definite) kernel. And the regularization term $\|f\|_K^2 = \mathbf{f}^T K \mathbf{f} = \sum_{k=1}^{V} \theta^k \|f\|_{K^{(k)}}^2$.

Similarly we define multiview Hessian. Suppose $H^j, j = 1,2,\cdots,V$ is the Hessian of the $j^{th}$ view, we have

$$H = \sum_{j=1}^{V} \beta^j H^j, s.t. \sum_{j=1}^{V} \beta^j = 1, \beta^j \geq 0, j = 1,2,\cdots,V.$$

According to the computation procedure of Hessian, $H^j$ is semi-definite positive. Then $H$ is also semi-definite positive. And we have the regularization term $\|f\|_I^2 = \mathbf{f}^T H \mathbf{f} = \sum_{j=1}^{V} \beta^j \|f\|_{I^{(j)}}^2$.

Substitute logistic loss function and the above regularizations into (1), the mHLR can be rewritten as:

$$\min_{f\in\mathcal{H}_K,\theta\in R^V,\beta\in R^V} -\frac{1}{l}\sum_{i=1}^{l}\left(y_i\log\frac{1}{1+e^{-f(x_i)}}+(1-y_i)\log\left(1-\frac{1}{1+e^{-f(x_i)}}\right)\right)$$

$$+\gamma_K\sum_{k=1}^{V}\theta^k\|f\|_{K^{(k)}}^2+\gamma_I\sum_{j=1}^{V}\beta^j\|f\|_{I^{(j)}}^2+\gamma_\theta\|\theta\|_2^2+\gamma_\beta\|\beta\|_2^2$$

$$s.t.\ \sum_{k=1}^{V}\theta^k=1,\theta^k\geq 0, k=1,2,\cdots,V,$$

$$\sum_{j=1}^{V}\beta^j=1,\beta^j\geq 0, j=1,2,\cdots,V \quad (2)$$

where $\|\theta\|_2^2$ and $\|\beta\|_2^2$ are regularization terms to avoid the model parameter overfitting to only one view, and $\gamma_\theta\in R^+$ and $\gamma_\beta\in R^+$ are the parameters to balance the contribution of the terms $\|\theta\|_2^2$ and $\|\beta\|_2^2$ respectively.

**Representer Theorem**: With fixed $\theta$ and $\beta$, the solution of (2) w.r.t $f$ exists and has the following expression

$$f^* = \sum_{i=1}^{l+u}\alpha_i K(x_i,x) \quad (3)$$

which is an expansion in terms of the labeled and unlabeled example.

The representer theorem shows the general form of the solution (2) when $\theta$ and $\beta$ are fixed. Substitute (3) into (2), we finally construct the mHLR as the following optimization problem

$$\min_{\alpha\in R^{l+u},\theta\in R^V,\beta\in R^V} -\frac{1}{l}\sum_{i=1}^{l}\left(y_i\log\frac{1}{1+e^{-K(x_i,x)\alpha}}+(1-y_i)\log\left(1-\frac{1}{1+e^{-K(x_i,x)\alpha}}\right)\right)$$

$$+\gamma_K\alpha^T K\alpha+\gamma_I\alpha^T KHK\alpha+\gamma_\theta\|\theta\|_2^2+\gamma_\beta\|\beta\|_2^2$$

$$s.t.\ \sum_{k=1}^{V}\theta^k=1,\theta^k\geq 0, k=1,2,\cdots,V,$$

$$\sum_{j=1}^{V}\beta^j=1,\beta^j\geq 0, j=1,2,\cdots,V \quad (4)$$

where $K=\sum_{k=1}^{V}\theta^k K^k$ and $H=\sum_{j=1}^{V}\beta^j H^j$.

In this paper, we use alternating optimization [22] to solve the problem (4) iteratively. First, fix $\theta$ and $\beta$ to solve $\alpha$. Then fix $\alpha$ and $\beta$ to solve $\theta$. Finally fix $\alpha$ and $\theta$ to solve $\beta$. In the following section we will detailed the optimization procedure.

3. Algorithms

In this section, we present the implementation of alternating optimization of problem (4). First, solve $\alpha$ with fixed $\theta$ and $\beta$. Then solve $\theta$ with fixed $\alpha$ and $\beta$. Finally solve $\beta$ with fixed $\alpha$ and $\theta$.

Given fixed $\theta$ and $\beta$, (4) can be expressed as follows

$$\min_{\alpha \in \mathbb{R}^{l+u}} -\frac{1}{l}\sum_{i=1}^{l}\left(y_i \log \frac{1}{1+e^{-K(x_i,x)\alpha}} + (1-y_i)\log\left(1-\frac{1}{1+e^{-K(x_i,x)\alpha}}\right)\right) + \gamma_A \alpha^T K \alpha + \gamma_I \alpha^T K H K \alpha \quad (5)$$

where $K = \sum_{k=1}^{V} \theta^k K^k$ and $H = \sum_{j=1}^{V} \beta^j H^j$.

The logistic loss function is differentiable, hence we can employ the conjugate gradient algorithm to optimize problem (5). The gradient of the objective function in (5) can be written as:

$$\nabla f(\alpha) = -\frac{\log e}{l}\sum_{i=1}^{l}\left(y_i K(x_i,x)e^{-K(x_i,x)\alpha}\left(\frac{1}{1+e^{-K(x_i,x)\alpha}}\right) + (1-y_i)(-K(x_i,x))\left(\frac{1}{1+e^{-K(x_i,x)\alpha}}\right)\right)$$
$$+ \gamma_A(K+K^T)\alpha + \gamma_I(KHK+KHK^T)\alpha$$

Then we have the efficient optimization procedure of conjugate gradient algorithm as below:

Step 1: initialize $\alpha^0 \in \mathbb{R}^{l+u}, \delta, d^0 = -\nabla f(\alpha^0), 0 < \varepsilon \ll 1, m = 0$.

Step 2: while $|f(\alpha^{m+1}) - f(\alpha^m)| > \varepsilon$, do

$$\alpha^{m+1} = \alpha^m + \delta d^m$$

$$d^{m+1} = -\nabla f(\alpha^{m+1}) + \frac{\|\nabla f(\alpha^{m-1})\|^2}{\|\nabla f(\alpha^m)\|^2} d^m$$

$$m = m + 1.$$

Step 3: $\alpha^* = \alpha^{m+1}$.

Given fixed $\alpha$ and $\beta$, problem (4) w.r.t. $\theta$ can be expressed as the following subproblem (6):

$$\min_{\theta} -\frac{1}{l}\sum_{i=1}^{l}\left(y_i \log\frac{1}{1+e^{-\sum_{k=1}^{V}\theta^k K^k(x_i,x)\alpha}} + (1-y_i)\log\left(1-\frac{1}{1+e^{-\sum_{k=1}^{V}\theta^k K^k(x_i,x)\alpha}}\right)\right)$$
$$+\gamma_N\alpha^T\left(\sum_{k=1}^{V}\theta^k K^k(x_i,x)\right)\alpha + \gamma_r\alpha^T\left(\sum_{k=1}^{V}\theta^k K^k(x_i,x)\right)H\left(\sum_{k=1}^{V}\theta^k K^k(x_i,x)\right)\alpha$$
$$+\gamma_\theta\|\theta\|_2^2$$

$$s.t.\ \sum_{k=1}^{V}\theta^k = 1, \theta^k \geq 0, k = 1,2,\cdots,V \qquad (6)$$

where $H = \sum_{i=1}^{V}\beta^i H^i$.

The gradient of the objective function in (6) w.r.t. $\theta^k$ can be written as:

$$\nabla f(\theta^k) = \frac{\log e}{l}\sum_{i=1}^{l}\left(K^k(x_i,x)\alpha\left(y_i - \frac{1}{1+e^{-\sum_{k=1}^{V}\theta^k K^k(x_i,x)\alpha}}\right)\right) + \gamma_N\alpha^T K^k \alpha$$
$$+ \gamma_r\left(2H\left(\sum_{k=1}^{V}\theta^k K^k(x_i,x)\right)\alpha\right)^T K^k\alpha + \gamma_\theta\theta^k$$

Then we similarly have the conjugate gradient algorithm to optimize $\theta$ as follows.

Step 1: initialize $\theta^{(0)} \subset R^V, \delta, 0 < \varepsilon \ll 1, m = 0, d^{k(0)} = \nabla f\left(\theta^{k(0)}\right), k = 1,2,\cdots,V$.

Step 2: while $\left|f\left(\theta^{(m+1)}\right) - f\left(\theta^{(m)}\right)\right| > \varepsilon$, do

$$\begin{cases} \theta^{k(m+1)} = \theta^{k(m)} + \delta d^{k(m)} \\ d^{k(m+1)} = -\nabla f\left(\theta^{k(m-1)}\right) + \frac{\left|\nabla f\left(\theta^{k(m-1)}\right)\right|^2}{\left\|\nabla f\left(\theta^{k(m)}\right)\right\|^2}d^{k(m)}, k = 1,2,\cdots,V \end{cases}$$

$m = m + 1$.

Step 3: $\theta^* = \theta^{(m+1)}$.

Given fixed $\alpha$ and $\theta$, we have the subproblem (7) of (4) w.r.t. $\beta$.

$$\min_{\beta \in R^V} \gamma_i \alpha^T K \left( \sum_{j=1}^{V} \beta^j H^j \right) K\alpha + \gamma_\beta \|\beta\|_2^2$$

$$s.t. \sum_{j=1}^{V} \beta^j = 1, \beta^j \geq 0, j = 1,2,\cdots,V \qquad (7)$$

where $K = \sum_{k=1}^{V} \theta^k K^k$. Subproblem (7) can be viewed as the learning of the optimal linear combination of multiple Hessians.

By using alternating optimization algorithm, we obtain a local optimal solution of problem (4).

4. Experiments

To evaluate the effectiveness of the proposed mHLR, we conduct action recognition experiments on the unstructured social activity attribute (USAA) dataset [19] [20] which is a subset of CCV dataset [23] . This dataset contains about 1,600 videos from 8 different semantic class videos which are home videos of social occasions including birthday party, graduation party, music performance, non-music performance, parade, wedding ceremony, wedding dance and wedding reception.

In our experiments, we use the 69 ground-truth attributes provided by Fu et al. [19] [20]  as tagging features and low features including SIFT, STIP and MFCC according to [23]   as visual features (Note that several videos are excluded for there low-level features are all zeros due to some problems of extracting process).

We use around 100 videos per-class for training and testing respectively. In the semi-supervised learning experiments, we assign 10%, 30%, 50%, 70% and 100% examples in training set as labeled data and the rest as unlabelled data. In particular, parameters $\gamma_K, \gamma_l$ are tuned from the candidate set $\{10^e | e = -8, -7, \cdots, 1, 2\}$ and $\gamma_\theta$ and $\gamma_\beta$ are both set to 0.1. The number of the neighbors in computing Hessian is set to 100 and the number in computing graph Laplacian is set to 50 for all experiments.

We compare the proposed mHLR with multiview Laplacian regularized logistic repression (mLLR) and the feature concatenation method (mCLR) that concatenates the visual feature and tagging feature into one feature vector). We also compare mHLR with the related single view methods. We use the average precision (AP) [24] for each class and mean average precision (mAP) for all classes as measure criteria.

Figure 1 is the mAP of different methods including logistic regression on visual feature (VisF), Laplacian regularized logistic regression on visual feature (LapVF), Hessian regularized logistic regression on visual feature (HesVF), logistic regression on tagging feature (TagF), Laplacian regularized logistic regression on tagging feature (LapTag), Hessian regularized logistic regression on tagging feature (HesTag), logistic regression on multiview concatenation feature (mCLR), multiview Laplacian regularized logistic regression (mLLR) and multiview Hessian regularized logistic regression (mHLR). The x-axis is the number of labeled examples. From Figure 2, we can see that Hessian regularization performs better than Laplacian regularization in both single view and multivew cases. The smaller the labeled example number is, the more significant the performance improvement is. And Figure 2 also shows that multiview methods outperform the related single view methods.

Figure 2 is the AP of different multiview methods including mCLR, mLLR and mHLR on some selected human action classes. Each subfigure corresponds to one human action of the USAA dataset. The x-axis is the number of labeled examples. From Figure 2, we can see that, multiview manifold regularization based logistic regression methods significant improve the performance comparing with the concatenation method especially when there is a small number of labeled examples. And mHLR outperforms mLLR in most cases.

5. Conclusion

Manifold regularized semi-supervised learning (SSL) has been successfully applied to many computer vision applications including image classification and video annotation. However, most existing algorithms are not applicable to the practical human action videos which are often represented by multi-modal content and different representations. Therefore, in this paper we propose multiview Hessian regularized logistic regression (mHLR) for human action recognition. With the help of Hessian regularization and logistic loss function, mHLR (1) combines multiple Hessian regularizations which significantly boost the exploring of local geometry; (2) naturally handles multi-view instances with multiple representations; (3) has efficient and effective optimization algorithm based on the smooth loss function. Extensive experiments are conducted on the unstructured social activity attribute (USAA) dataset and the experimental results validate the effectiveness of the proposed mHLR for human action recognition comparing with other related algorithms including mLLR, mCLR.

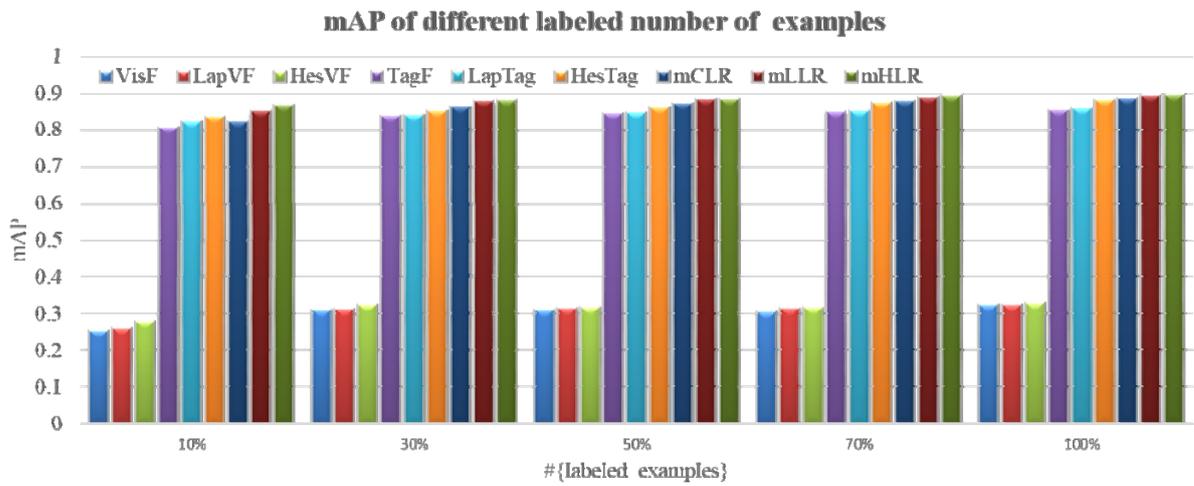

Figure 1. mAP of different methods, the methods from left to right are VisF, LapVF, HesVF, TagF, LapTag, HesTag, mCLR, mLLR, mHLR.

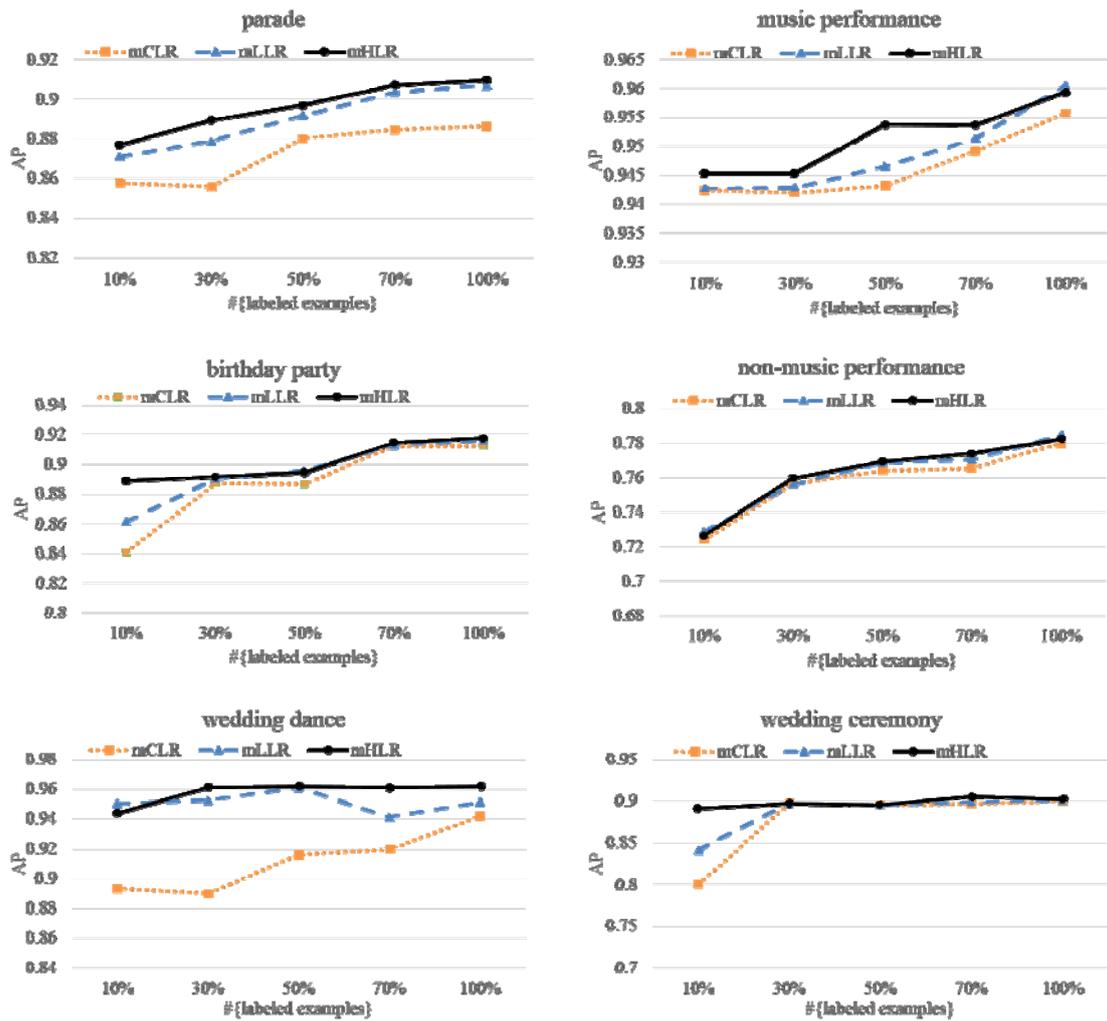

Figure 2. AP of different multiview methods on some selected action classes including parade, music performance, birthday party, non-music performance, wedding dance and wedding ceremony.